# Explaining Adverse Actions in Credit Decisions Using Shapley Decomposition

Vijayan N. Nair, Tianshu Feng, Linwei Hu,
Zach Zhang, Jie Chen, and Agus Sudjianto

Corporate Model Risk, Wells Fargo
April 25, 2022

**Abstract**

When a financial institution declines an application for credit, an adverse action (AA) is said to occur. The applicant is then entitled to an explanation for the negative decision. This paper focuses on credit decisions based on a predictive model for probability of default and proposes a methodology for AA explanation. The problem involves identifying the important predictors responsible for the negative decision and is straightforward when the underlying model is additive. However, it becomes non-trivial even for linear models with interactions. We consider models with low-order interactions and develop a simple and intuitive approach based on first principles. We then show how the methodology generalizes to the well-known Shapely decomposition and the recently proposed concept of Baseline Shapley (B-Shap). Unlike other Shapley techniques in the literature for local interpretability of machine learning results, B-Shap is computationally tractable since it involves just function evaluations. An illustrative case study is used to demonstrate the usefulness of the method. The paper also discusses situations with highly correlated predictors and desirable properties of fitted models in the credit-lending context, such as monotonicity and continuity.

## 1. Introduction

In credit lending, an **adverse action** (AA) occurs when a financial institution declines an application for credit. This covers several situations: i) a refusal to grant credit in substantially the amount or terms requested in an application, ii) a termination of an account or an unfavorable change in the terms of an account; or iii) a refusal to increase the amount of credit available to an applicant. The institution is then required to provide clear reasons for the AA or to let the applicant know that they are entitled to request an explanation (Regulation B of the Equal Credit Opportunity Act, www.federalreserve.gov/boarddocs/supmanual/cch/fair_lend_reg_b.pdf; and Fair Credit Reporting Act, www.consumerfinance.gov/compliance/compliance-resources/other-applicable-requirements/fair-credit-reporting-act/).

These regulations are not prescriptive in terms of the explanations except that they must be clear and well supported. In practice, the reasoning can be simple if the application is incomplete or has unverifiable information. Even when a negative decision is based on a full review of the application, the explanation will depend on whether the process is based on judgment, predictive model, or a combination of the two. In this paper, we consider situations where the decision is based statistical or machine learning algorithms. Specifically, we will focus on models for probability of default (denoted as **PoD** from now on).

Financial institutions often use historical data from their current or past customers to develop their own model. To describe this, let $\{y_i, x_i\}, i = 1, \ldots n$, be the data, where i) $y_i$ is a binary response (1 or 0) denoting whether the $i$–th customer defaulted or not over some time period (say within the last year); and ii) $\{x_i = (x_{1i}, \ldots, x_{Ki})\}$'s are $K$ variables (predictors, features) associated with the customers at the time of the loan application. Using the data, the institution develops its own predictive model, $p(x)$, for PoD. Given a new applicant with attribute $x^*$, the predicted PoD, $p(x^*)$, is used to make accept/decline decision. More specifically: i) a threshold $\tau$ is first selected; and ii) a new loan application is accepted or declined depending on whether $p(x^*) \leq \tau$ or $> \tau$. It is important to keep in mind, however, that the decision process is usually more complex and may be made by a human-in-the-loop who is guided by the model results.

Section 2 formulates the AA explanation problem as follows. Given a declined application with attribute $x^D$, one picks a reference point $x^A$ in the "acceptance" region and decomposes the difference in the PoDs, $[p(x^D) - p(x^A)]$, and allocates them to the individual predictors $(x_1, \ldots, x_K)$. It is preferable to consider the differences in the log-odds or logit space since the models are typically fitted using this link function. Section 2.2 briefly discusses choices of reference points $x^A$ in the acceptance region.

Section 3 motivates our method for AA explanation from first principles for models with two- and three-factor interactions. The idea is intuitive and provides a natural way to decompose the difference of a function at two points to its constituent components. Further, the computations are simple for models with lower-order interactions. We provide explicit expressions for 2- and 3-order models.

The general problem is considered in Section 4 where we show that our approach is a special case of the well-known Shapley decomposition and Baseline Shapley (B-Shap) of Sundararajan and Najmi (2020). They proposed it as an alternative to SHAP explanations for local diagnostics (Lundberg and Lee, 2017). But its adaptation to AA explanation is natural and straightforward. Like all Shapley-based methods, B-Shap is model (or algorithm) agnostic technique. Unlike local SHAP which requires integrations, B-Shap is computationally tractable since it depends on just function evaluations.

Section 5 demonstrates the usefulness of the results using an illustrative case study. We describe the dataset, use a monotone neural network (Mono NN) algorithm to fit the model, compare its performance with an unconstrained feedforward NN, and demonstrate AA explanation for results based on Mono NN.

Section 6 discusses requirements of shape constraints such as monotonicity and continuity in the fitted model. The paper concludes with some remarks.

There are several approaches in the literature for explaining the fitted models using machine-learning algorithms. Some of these, such as Local Interpretable Model-Agnostic Explanations or LIME (Ribeiro, Singh, & Guestrin, 2016), involve fitting a local linear model at a point of interest and using the coefficients of the fitted model to explain the result locally. These techniques will not work for the AA explanation problem when the declined and accepted points, $x^D$ and $x^A$, are far apart and a local linear model does not provide a good fit.

## 2. Problem Formulation
### 2.1 Background

Let $p(x)$ be the predicted PoD of an applicant with predictor $x = (x_1, \dots x_K)$, obtained by fitting a model to historical data. Let $\tau$ be the suitably chosen threshold. The credit-decision algorithm will approve a future loan application with attribute $x^*$ if $p(x^*) \leq \tau$ and decline otherwise. In practice, $p(x)$ is often developed in terms of a link function such as

$$f(x) = logit[p(x)]. \qquad Eq(1)$$

Since the model is likely to be simpler or more interpretable in terms of $f(x)$, we will develop explanation in terms of the differences in $f(x)$ rather than $p(x)$. However, the arguments will also work with $p(x)$.

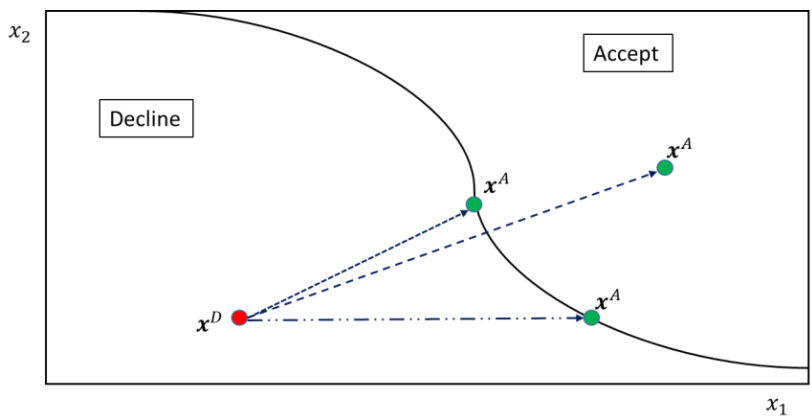

Figure 1: Selection of a reference point for comparison

The AA explanation involves providing information to the applicant why their loan is declined. A common approach is to compare their loan characteristic, $x^D$, with $x^A$, those of another applicant whose loan would be approved (Figure 1). Specifically:

a) For each customer who was declined, one selects a reference point for comparison in the "acceptance" region of the decision space. It can be the same point for all declined applicants, or it can depend on the particular declined applicant;
b) Decompose the difference in default probabilities between the declined and accepted applicants and attribute them to differences in the predictors.

Mathematically, this involves decomposing the difference as

$$[f(x^D) - f(x^A)] = \sum_k E_k(x^D, x^A), \qquad Eq\ (2)$$

where $E_k(x^D, x^A)$ is the contribution of the $k$-th predictor to the difference. This will allow the institution to attribute the AA to important predictors. In the rest of the paper, we denote $E_k(x^D, x^A)$ as simply $E_k$.

## 2.2 Selecting a reference point for comparison

Consider the illustrative diagram in Figure 1 that portrays the decision boundary separating regions where loans are declined from where loans are accepted. Here $x^D = (x_1^D, x_2^D)$ denotes the loan application that has been declined. We briefly consider a few options for selecting a reference point for comparison in the "accept" region. In practice, this choice will depend also on many other practical considerations. We discuss it here only to provide some concreteness. Any reference point will work for our methodology.

a) A common practice is to choose the reference point $x^A$ at some high values of the predictors in the acceptance region, typically maximum or close to maximum values. Arguably, a more reasonable choice would be some percentiles (such as median) of these predictors.
b) From the loan applicant's perspective, the most favorable point $x^A$ is the one with the shortest distance from $x^D$ to the decision boundary. This would provide the applicant with information on the smallest changes they have to make to be successful (sometimes called counterfactuals). There are, however, some potential challenges with this approach: i) the decision boundary is estimated from data and one should also consider the inherent variability; and ii) the reference point will vary with the declined loan and hence may be challenging to explain to multiple customers.
c) The above choices assume that the loan applicant will be able to improve their credit worthiness equally well for the different attributes. In practice, they may be able to manipulate a subset more easily than others. In this case, one can consider reference points that have the shortest distance in a lower dimensional subspace of the attributes. Figure 1 shows this reference point in terms of the shortest distance in $x_1$-dimension.

In the rest of this paper, we treat the reference point generically and do not consider any particular choice. We will use the following notation:
- $x^D = (x_1^D, \ldots x_K^D)$ denotes the point, and values of the predictors, for the declined application; and
- $x^A = (x_1^A, \ldots x_K^A)$ denotes the reference point, and values of the predictors, for the successful application that has been selected for comparison. This point can depend on $x^D$ but we will suppress any dependence.

## 3. AA Explanation in Important Special Cases

We start with special situations and develop the explanation method from first principles.

### 3.1 Additive Models with No Interactions

Suppose the fitted model is linear without interactions:

$$f(x) = b_0 + b_1 x_1 + \cdots + b_K x_K.$$

The decomposition in Eq (2) is then easy, with $E_k = b_k(x_k^D - x_k^A)$, for $k = 1, \ldots, K$.

Suppose now we have $f(x) = g_1(x_1) + \cdots + g_K(x_K)$, a generalized additive model (GAM). Then

$$f(x^D) - f(x^A) = [g_1(x_1^D) - g_1(x_1^A)] + \cdots + [g_K(x_K^D) - g_K(x_K^A)],$$

so the $E_k$'s are again easily computed.

### 3.2 Two-Factor Linear Model with Interaction

The explanations become less obvious in the presence of even simple interactions. Consider a two-factor linear model with interaction: $f(x) = b_0 + b_1 x_1 + b_2 x_2 + b_{12} x_1 x_2$. Now,

$$[f(x^D) - f(x^A)] = b_1(x_1^D - x_1^A) + b_2(x_2^D - x_2^A) + b_{12}(x_1^D x_2^D - x_1^A x_2^A). \quad Eq\ (3)$$

The difference in interactions, last term on the right-hand side in Eq (3), involves both $x_1$ and $x_2$ and has to be further decomposed and allocated to each separately. Before providing a solution to this special case, we consider a more general model with two factors in the next section.

### 3.3 General Model with Two Factors

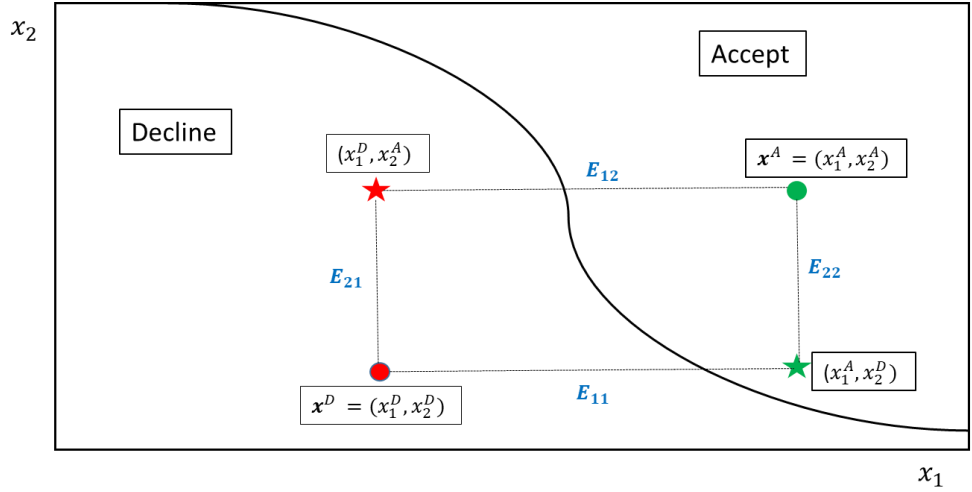

Figure 2: Graphical illustration of the different components in Eq (4) and Eq (5)

Suppose we have fitted a general model of two variables $f(x_1, x_2)$. Consider the rectangle in Figure 2 and the points labelled $E_{ij}$, $i, j = 1,2$. Define $E_{11} = [f(x_1^D, x_2^D) - f(x_1^A, x_2^D)]$ and the other terms similarly. Note first that

$$[f(x_1^D, x_2^D) - f(x_1^A, x_2^A)] = E_{11} + E_{22} = E_{21} + E_{22} = \frac{1}{2}(E_{11} + E_{12}) + \frac{1}{2}(E_{21} + E_{22}).$$

Further, $E_{11} = [f(x_1^D, x_2^D) - f(x_1^A, x_2^D)]$ measures the difference when $x_1$ changes from its level at the declined point to its level at the accepted point, with $x_2$ is fixed at the declined point. Similarly, $E_{12} = [f(x_1^D, x_2^A) - f(x_1^A, x_2^A)]$ measures the difference with $x_2$ fixed at the accepted point. When there is no interaction between $x_1$ and $x_2$, $E_{11} = E_{12}$. Thus, it is natural to assign the contribution (or attribution) for $x_1$ as the average of these two values:

$$E_1 = \frac{1}{2}(E_{11} + E_{12}) = \frac{1}{2}\{[f(x_1^D, x_2^D) - f(x_1^A, x_2^D)] + [f(x_1^D, x_2^A) - f(x_1^A, x_2^A)]\}, \quad Eq\ (4).$$

Similarly,

$$E_2 = \frac{1}{2}(E_{21} + E_{22}) = \frac{1}{2}\{[f(x_1^D, x_2^D) - f(x_1^D, x_2^A)] + [f(x_1^A, x_2^D) - f(x_1^A, x_2^A)]\}. \quad Eq\ (5)$$

To compute the attributions, we have to do only four function evaluations (at the four corners of the rectangle in Figure 2) and compute the differences in Eq (4) and Eq (5).

### 3.4 Revisiting the Simple Case in Section 3.2

Applying Equations (4 – 5) to the interaction term in Eq (3), we get

$$b_{12}(x_1^D x_2^D - x_1^A x_2^A) = \frac{1}{2}[b_{12}(x_1^D x_2^D - x_1^A x_2^D) + b_{12}(x_1^D x_2^A - x_1^A x_2^A)]$$
$$+ \frac{1}{2}[b_{12}(x_1^D x_2^D - x_1^D x_2^A) + b_{12}(x_1^A x_2^D - x_1^A x_2^A)].$$

Simplifying,

$$b_{12}(x_1^D x_2^D - x_1^A x_2^A) = \left[b_{12}(x_1^D - x_1^A)\frac{x_2^D + x_2^A}{2}\right] + \left[b_{12}(x_2^D - x_2^A)\frac{x_1^D + x_1^A}{2}\right]. \quad Eq\ (6)$$

Note the symmetry in the decomposition on the right-hand side in Eq (6). Combining the main effects in Section 3.2 with the interactions, we get the decompositions for the two factors as

$$E_1 = b_1(x_1^D - x_1^A) + \frac{1}{2}b_{12}(x_1^D - x_1^A)(x_2^D + x_2^A),$$

and

$$E_2 = b_2(x_2^D - x_2^A) + \frac{1}{2}b_{12}(x_2^D - x_2^A)(x_1^D + x_1^A).$$

### 3.5 $K$ factors with Second Order Terms Only

The results in Section 3.4 generalize easily to $K$ factors. To ease into the general case, suppose first $K = 3$. Then,

$$f(x_1, x_2, x_3) = f_{12}(x_1, x_2) + f_{13}(x_1, x_3) + f_{23}(x_2, x_3). \quad Eq(7)$$

We can apply the decomposition in Section 3.3 to each of the three pairwise interactions. For example,

$$E_1 = \frac{1}{2}\{[f_{12}(x_1^D, x_2^D) - f_{12}(x_1^A, x_2^D)] + [f_{12}(x_1^D, x_2^A) - f_{12}(x_1^A, x_2^A)]\} +$$
$$\frac{1}{2}\{[f_{13}(x_1^D, x_3^D) - f_{13}(x_1^A, x_3^D)] + [f_{13}(x_1^D, x_3^A) - f_{13}(x_1^A, x_3^A)]\}.$$

This extends in a straightforward manner to models with $K$ predictors of the form

$$f(x_1, \ldots, x_K) = \sum_{i \neq j} f_{ij}(x_i, x_j). \quad Eq(8)$$

The attribution for $x_k$ ($k$ is fixed in the summation below) is:

$$E_k = \sum_{j \neq k} \frac{1}{2}\{[f_{kj}(x_k^D, x_j^D) - f_{kj}(x_k^A, x_j^D)] + [f_{kj}(x_k^D, x_j^A) - f_{kj}(x_k^A, x_j^A)]\}. \quad Eq\ (9)$$

To compute the decompositions. We have to do a total of $(K\ choose\ 2) \times 4$ function evaluations: $(K\ choose\ 2)$ sub-models of two-factors, and each requiring four evaluations.

## 3.6 Two-factor Models Expressed as Orthogonal Main Effects and Interactions

The $K$-factor interaction model in Eq (8) can be re-expressed as main effects (GAM) and two-factor functional interactions that are constrained to be orthogonal:

$$f(x_1, \ldots, x_K) = \sum_j g_j(x_j) + \sum_{i \neq j} g_{ij}(x_i, x_j).$$

These have been called GA2M or GAMI in the literature. We can express the AA explanations in Eq (9) in terms of main effects and interactions as

$$E_k = [g_k(x_k^D) - g_k(x_k^A)]$$
$$+ \sum_{j \neq k} \frac{1}{2}\{[g_{kj}(x_k^D, x_j^D) - g_{kj}(x_k^A, x_j^D)] + [g_{kj}(x_k^D, x_j^A) - g_{kj}(x_k^A, x_j^A)]\}.$$

## 3.7 Model with Three Factors

Suppose $K = 3$ with $f(x) = f(x_1, x_2, x_3)$. Consider the path in Figure 3 from $x^D$ to $x^A$: $(x_1^D, x_2^D, x_3^D) \to (x_1^A, x_2^D, x_3^D) \to (x_1^A, x_2^A, x_3^D) \to (x_1^A, x_2^A, x_3^A)$. For this path,

$$[f(x^D) - f(x^A)] = [f(x_1^D, x_2^D, x_3^D) - f(x_1^A, x_2^D, x_3^D)] + [f(x_1^A, x_2^D, x_3^D) - f(x_1^A, x_2^A, x_3^D)] +$$
$$[f(x_1^A, x_2^A, x_3^D) - f((x_1^A, x_2^A, x_3^A)]. \qquad Eq\ (10)$$

There are six such paths in Figure 3, and $[f(x^D) - f(x^A)]$ can be written in terms of the function evaluations of the points on those paths, as in Eq (10). We can also write the difference on the left hand-side as an average of the expressions for the six paths.

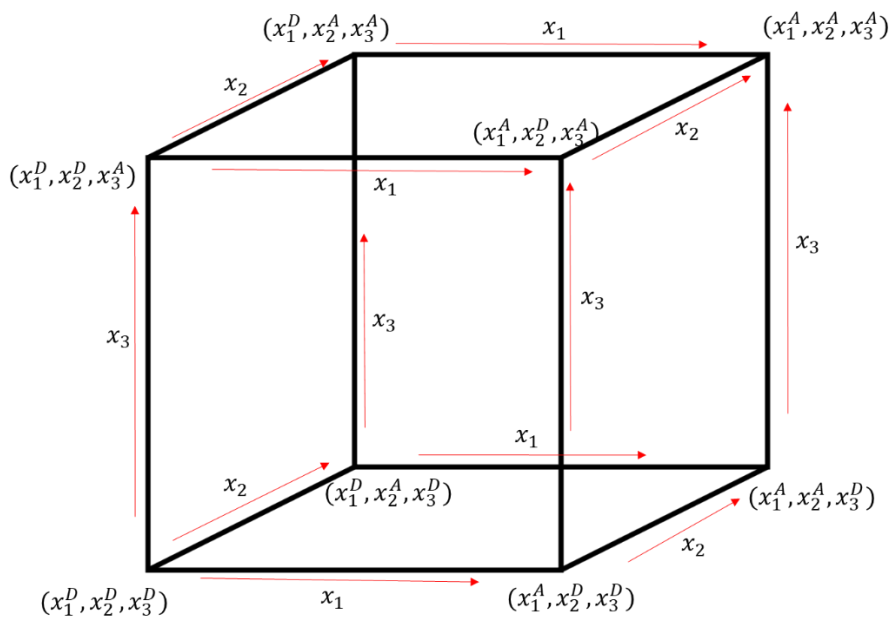

Figure 3: Graphical illustration for a three-factor model

To get the contribution $E_1$ for $x_1$, we collect all the terms involving differences in $x_1^D$ and $x_1^A$ with the settings of $(x_1, x_2)$ fixed. Take the term in Eq (10) is an example. Rearranging terms,

$$E_1 = \frac{1}{3}[f(x_1^D, x_2^D, x_3^D) - f(x_1^A, x_2^D, x_3^D)] + \frac{1}{6}\{[f(x_1^D, x_2^A, x_3^D) - f(x_1^A, x_2^A, x_3^D)] + [f(x_1^D, x_2^D, x_3^A) - f(x_1^A, x_2^D, x_3^A)]\} + \frac{1}{3}[f(x_1^D, x_2^A, x_3^A) - f(x_1^A, x_2^A, x_3^A)].$$

Using similar arguments,

$$E_2 = \frac{1}{3}[f(x_1^D, x_2^D, x_3^D) - f(x_1^D, x_2^A, x_3^D)] + \frac{1}{6}\{[f(x_1^D, x_2^D, x_3^A) - f(x_1^D, x_2^A, x_3^A)] + [f(x_1^A, x_2^D, x_3^D) - f(x_1^A, x_2^A, x_3^D)]\} + \frac{1}{3}[f(x_1^A, x_2^D, x_3^A) - f(x_1^A, x_2^A, x_3^A)],$$

and

$$E_3 = \frac{1}{3}[f(x_1^D, x_2^D, x_3^D) - f(x_1^D, x_2^D, x_3^A)] + \frac{1}{6}\{[f(x_1^A, x_2^D, x_3^D) - f(x_1^A, x_2^D, x_3^A)] + [f(x_1^D, x_2^A, x_3^D) - f(x_1^D, x_2^A, x_3^A)]\} + \frac{1}{3}[f(x_1^A, x_2^A, x_3^D) - f(x_1^A, x_2^A, x_3^A)].$$

The computation in this case require function evaluations at $2^3 = 8$ corner points of the cube in Figure 3.

### 3.8 General Models with Only Three-Order Terms

Consider the case with K factors and suppose the model depends on three-order submodels only:

$$f(x_1, \ldots, x_K) = \sum_{i \neq j \neq k} f_{ijk}(x_i, x_j, x_k)$$

The results in the last section can be generalized in a straignforward way to get the attribution for $x_\ell$, the $\ell$ −th variable, as the sum across $(j, k)$ with $j \neq k$ and $\ell$ fixed:

$$E_\ell = \sum_{\ell \neq j \neq k} \frac{1}{3}[f(x_\ell^D, x_j^D, x_k^D) - f(x_\ell^A, x_j^D, x_k^D)]$$
$$+ \frac{1}{6}\{[f(x_\ell^D, x_j^A, x_k^D) - f(x_\ell^A, x_j^A, x_k^D)] + [f(x_\ell^D, x_j^D, x_k^A) - f(x_\ell^A, x_j^D, x_k^A)]\}$$
$$+ \frac{1}{3}[f(x_\ell^D, x_j^A, x_k^A) - f(x_\ell^A, x_j^A, x_k^A)].$$

Now, we have a total of ($K$ choose 3) × 8 function evaluations: ($K$ choose 3) sub-models of three factors with each requiring 8 function evaluations.

## 4. General Case

So far, we have developed AA explanation for special cases from first principles in an intuitive manner. We now consider the general case of an arbitrary model with $K$ factors, and show that our results are special cases of Shapley decomposition (Shapley, 1953) and Baseline Shapley (Sundararajan and Najmi, 2020).

### 4.1 Shapley Decomposition

Given an arbitrary function $f(x)$ of $K$ predictors, the goal is to decompose the difference $[f(x^D) - f(x^A)]$ into individual components of each predictor. This problem was addressed in Shapley (1953) in the context of cooperative games. His formulation was as follows. There are K players involved in a coalition in a cooperative game, and the goal is to distribute the gains from the game among the players. Let $\boldsymbol{K} = \{1, 2, \dots, K\}, \{k\}$ be the player of interest, $\boldsymbol{K}\backslash k$ denote a subset that does include $\{k\}$, $\boldsymbol{S}$ is any subset of $\boldsymbol{K}$, $|\boldsymbol{S}|$ is its cardinality, and $val(\boldsymbol{S})$ is some value function for the game that is associated with the set $\boldsymbol{S}$. Shapley (1953) showed that the unique distribution scheme that satisfies several desirable axioms is

$$\phi_k = \sum_{S \subseteq K\backslash k} \frac{|S|!\,(|K| - |S| - 1)!}{|K|!} \bigl(val(\boldsymbol{S} \cup k) - val(\boldsymbol{S})\bigr) \qquad Eq\ (11)$$

for $k = 1, \dots, K$. The summation in Eq (11) is over all possible subsets $\boldsymbol{S}$, and the combinatorial coefficients arise from the number of such subsets.

Owen (2014) proposed using Eq (11) with the value function $val(\boldsymbol{S}) = Var(\mathbb{E}(f(X)|X_S))$ as a measure of global variable importance in complex statistical and ML models. This is an alternative to the commonly used variable importance analysis based on permutation (Breiman, 2001). However, when the sample size and the number of variables $K$ (and hence the number of possible subsets) are large, computing the value function (variances of the conditional expectation) is challenging. This has impeded the use of Global Shapley for variable importance analysis.

Consider a different (local) explanation problem in ML algorithms. Given a fitted model $f(x)$ and a point of interest $x^*$ in the prediction space, the goal is to decompose $f(x^*)$ into contributions of the $K$ variables. Lundberg and Lee (2017 proposed using Eq (11) with the value function as $val(\boldsymbol{S}) = \mathbb{E}(f(X)|X_S)$ and referred to it SHapley Additive explanation (or SHAP). While this is simpler to compute than global Shapley, it is still time-consuming when $K$, and hence the number of sub-models, is large. This has led to the development of various approximations in the literature such as Kernel and Tree SHAP (Lundberg, Erion, and Lee, 2018), and the use of marginal expectations.

Sundararajan and Najmi (2020) proposed another alternative to local explanation called Baseline-Shapley (referred to as B-Shap). Instead of using integrations, it involves computing just the difference of the functions at each data point vs at an average reference point, and hance is easier to compute. We adapt in in a straightforward manner for AA explanation in the next section.

**4.2 Baseline-Shapley (B-Shap) for AA Explanation**

We use a slightly different notation in this section. Let $\boldsymbol{S_k} = \boldsymbol{S} \backslash \{k\}$, the subset of $\boldsymbol{S}$ without $\{k\}$. Then, $E_k$, the AA explanation for the $k$-th variable is:

$$E_k = \sum_{S_k \subseteq K \backslash \{k\}} \frac{|S_k|!\,(|K|-|S_k|)!}{|K|!} \left( f(x_k^D; x_{S_k}^D; x_{K \backslash S}^A) - f(x_k^A; x_{S_k}^D; x_{K \backslash S}^A) \right). \quad Eq\ (12)$$

This is just Eq (11) with an appropriate value function for our application. To make the idea concrete, consider a simple example with $\boldsymbol{K} = \{1,2,3,4,5\}$, take $\{k\} = \{1\}$, and $\boldsymbol{S} = \{1,2,3\}$. Then, $\boldsymbol{S_k} = \{2,3\}$, and $\boldsymbol{K} \backslash \boldsymbol{S} = \{4,5\}$. The right-hand side of Eq (12) is then

$$[f(x_1^D, x_2^D, x_3^D, x_4^A, x_5^A) - f(x_1^A, x_2^D, x_3^D, x_4^A, x_5^A)].$$

As noted earlier, B-Shap involves only function evaluations, so it is computationally less expensive. For a model with $K$-variables, there are $2^K$ function evaluations. When the model structure is simpler, as in the special cases in Section 3, the computations will be even less demanding.

**4.3 Highly Correlated Predictors**

Given the automated nature of feature selection and feature engineering in ML algorithms, modelers tend to include a large number of predictors in the model. So it is likely that some of the predictors in the model will have moderate to high correlation. If so, some values involved in computing the differences in Eq (12) may lie outside of the data envelope where the fitted model is unreliable. It would make sense then to treat the correlated predictors jointly in making attributions. To make this notion concrete, suppose we have three predictors $(x_1, x_2, x_3)$ and $(x_1, x_2)$ are strongly correlated. In this case, we can view $(x_1, x_2)$ as a single predictor $z_1$ and treat it as a two-dimensional attribution problem with predictors $(z_1, x_3)$. We take $z_1^D = (x_1^D, x_2^D)$ and $z_1^A = (x_1^A, x_2^A)$. We will see an illustration of this joint attribution to the illustrative example in Section 5. Of course there are other ways to treat this problem, including use of principal components or other dimension-reduction techniques.

**5. Illustrative Application**

We use an illustrative example to demonstrate the application of the results. We first have to develop the model for PoD from historical data. We use a simulated dataset that is intended to mimic application to credit cards. The predictors and their marginal distributions were obtained from credit bureau data. Their correlation structure as well as the input-output model were simulated, but they were intended to mimic real-world behavior. Readers interested in just the demonstration of AA explanation can skip the next two sub-sections and go directly to Section 5.3.

### 5.1. Description of Dataset

We simulated historical information on 50,000 current customers from the financial institution for a comparable credit product. The response was binary with $y = 1$ if account defaulted during an 18-month period and $y = 0$ otherwise. There was also information on 10 credit-related predictors at the time the customers applied for the product. The first two columns of Table 1 provide the variable names and descriptions. Subject-matter expertise suggested that the relationship between PoD and six of the 10 variables should be monotone. This information provided the third column of Table 1. For our analysis, we standardized *x1 = average card balance, x2 = credit age,* and *x4 = total balance*. We applied a log transformation to *x8 = total amount currently past due*. Finally, in order for the direction of monotonicity to be the same for all variables, we flipped the sign of *x2 = credit age*, and called it *credit age flip*.

Table 1: Description of Response and Predictors

| Variable Name | Description | Monotone in probability of default |
|---|---|---|
| **Response:** $y = $ default indicator | $y = 1$ if account defaulted during an 18-month period, and $y = 0$ if it did not default. | |
| **Predictors** | | |
| x1 = avg bal cards *std* | Average monthly debt standardized: amount owed by applicant) on all of their credit cards over last 12 months | N |
| x2 = credit age *std* | Age in months of first credit product standardized: first credit cards, auto-loans, or mortgage obtained by the applicant | Y = Decreasing |
| x3 = pct over 50 uti | Percentage of open credit products (accounts) with over 50% utilization | N |
| x4 = tot balance *std* | Total debt standardized: amount owed by applicant on all of their credit products (credit cards, auto-loans, mortgages, etc.) | N |
| x5 = uti open card | Percentage of open credit cards with over 50% utilization | N |
| x6 = num acc 30d past due 12 months | Number of non-mortgage credit-product accounts by the applicants that are 30 or more days delinquent within last 12 months (Delinquent means minimum monthly payment not made) | Y = Increasing |
| x7 = num acc 60d past due 6 months | Number of non-mortgage credit-product accounts by the applicants that are 30 or more days delinquent within last 6 months | Y = Increasing |

| x8 = tot amount currently past due *log* | Total debt standardized: amount owed by applicant on all of their credit products – credit cards, auto-loans, mortgages, etc. | Y = Increasing |
| --- | --- | --- |
| x9 = num credit inq 12 month | Number of credit inquiries in last 12 months. An inquiry occurs when the applicant's credit history is requested by a lender from the credit bureau. This occurs when a consumer applies for credit. | Y = Increasing |
| x10 = num credit card inq 24-month | Number of credit card inquiries in last 24 months. An inquiry occurs when the applicant's credit history is requested by a lender from the credit bureau. This occurs when a consumer applies for credit. | Y = Increasing |

The correlation structure of the variables is given in Figure 4, and we see strong correlation structure among subsets of variables. This is not surprising since x1 and x4 are both measures of balance, x3 and x5 are measures of utilization, x6-x8 are measures of delinquency (past due), and x9 and x10 are measures of number of inquiries. This is typical in credit applications where multiple dimensions of the same latent variable are collected and used in the model. We simulated the correlation structure to mimic this real world scenario.

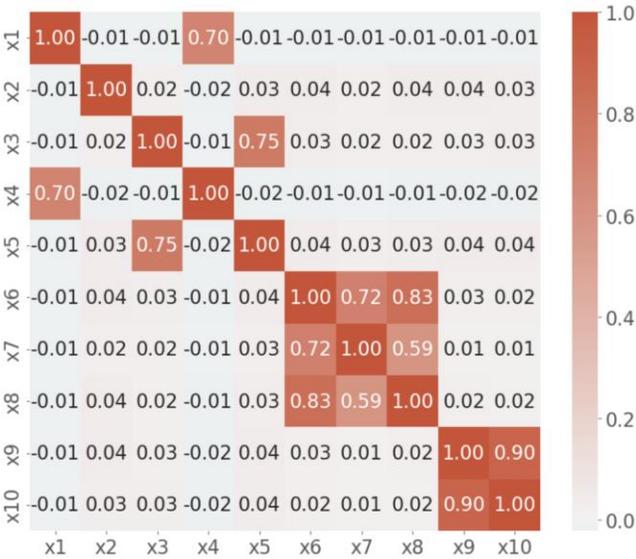

Figure 4: Correlation Matrix for the Predictors

**5.2. Model Development, Performance, and Diagnostics**

We fitted two different models: an unconstrained feedforward neural network (FFNN) and a monotone NN (mono-NN) that incorporates the shape constraints in Table 1. The mono-NN algorithm is based on the two-stage approach in Liu, Han, Zhang and Liu (2020).

The dataset with 50,000 observations was divided into training (80%), validation (10%), and testing (10%). We tuned the hyper-parameters of the FFNN algorithm and ended up with three layers and nodes [23, 35, 15] and learning rate of 0.004. The mono-NN also had three layers with nodes [35, 15, 5] and learning rate 0.001.

Table 2: Training and Test AUCs for the Two Algorithms

| Algorithm | Training AUC | Test AUC |
|---|---|---|
| FFNN | 0.810 | 0.787 |
| Mono-NN | 0.807 | 0.797 |

Table 2 shows the predictive performances. Mono-NN has a lower training AUC but higher test AUC, indicating it generalizes to the test dataset better. It also has a smaller gap between train and test AUCs, suggesting it might be more robust.

Figure 5 are variable importance plots for the mono-NN model. The left panel shows results for all 10 predictors. Credit age (x2) is by far the most important followed by x9 = number of credit inquiries in last 12 months. Note, however, that x10 = number of credit card inquiries in the last 12 months is the least important. This may be because x9 and x10 are highly correlated. Similar issues arise with x6, x7, and x8 (measures of delinquency or accounts past due), whose effects may be distributed. To address possible interpretation problems from high levels of correlation, we collapsed the 10 to get five predictors that measure intrinsically different quantities. The right panel in Figure 5 shows the joint variable importance. Credit age remains the most important variable, but now delinquency (accounts past due) and inquiries are showing up collectively as important. These results are more meaningful in this application.

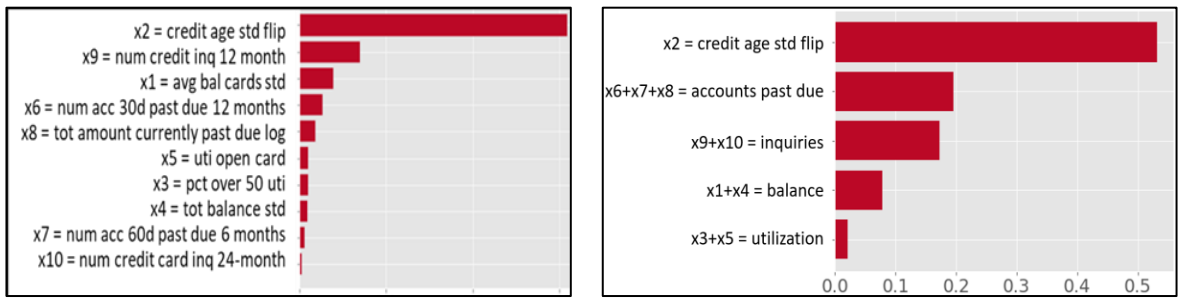

Figure 5. Permutation-based variable importance scores for Mono-NN. Left panel: importance values for all variables; Right panel: joint importance for the collapsed five variables

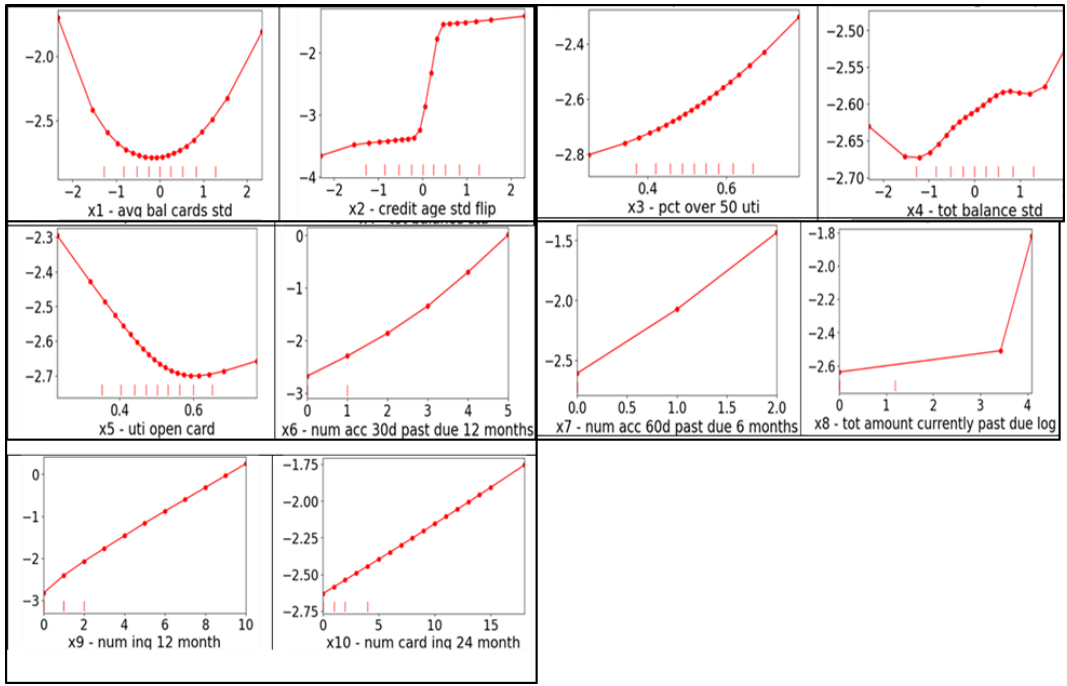

Figure 6. One-dimensional PDPs for the 10 variables based on Mono-NN algorithm

Figure 6 shows the partial-dependence plots (PDPs) describing one-dimensional input-output relationships: x-axis corresponds to the values of the variables and y-axis corresponds to the log-odds of PoD. Recall from Table 1 that the Mono-NN was constrained to be monotone in x2 and x6-x10, and the PDPs retain this shape constraint. In fact, x7, x9 and x10 are mostly linear. In addition, the shape for x3 is also monotone. The average cash balance = x1 and utilization = x5 have quadratic behavior, while total balance = x4 has a more complex pattern.

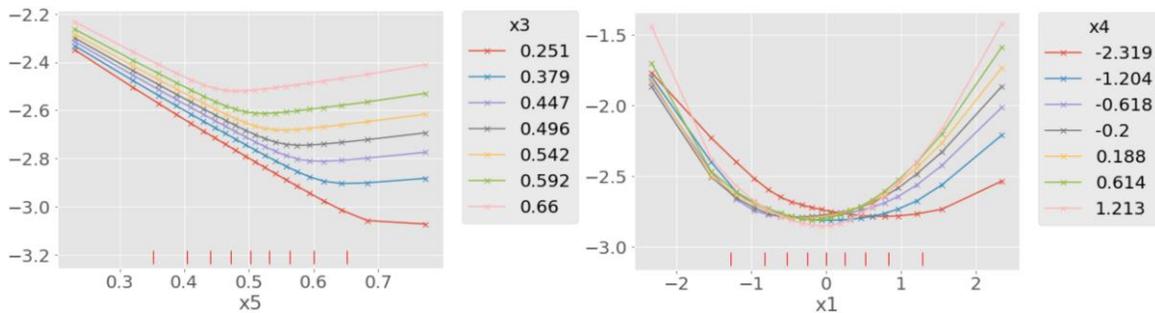

Figure 7. Two-Dimensional PDPs for Selected Predictors for Mono-NN

  We also examined the interactions among the predictors using H-statistics and two-dimensional PDPs (Friedman, 2001). Figure 7 shows the two pairs of variables for which there was some degree of interaction. The plots shown are slices of one-dimensional PDPs for the variables on the x-axis with the colored curves corresponding to different slices of the other variable. If there is no interaction, the curves will be parallel. The left panel shows interaction between x3 (percent of all open credit products with over 50% utilization) and x5 (percentage of open credit cards with over 50% utilization). They are both measuring credit utilization. The right panel shows interaction between x1 (average monthly debt) and x4 (total debt standardized). These two are both measuring debt.

### 5.3. Adverse Action Explanation Using B-Shap

Finally, we turn to the demonstration of AA explanation in this example. For illustrative purposes, we chose the PoD threshold as $\tau = 0.25$. In other words, any future credit card application with $p(x^*) \leq 0.25$ will be approved; otherwise it will be declined.

The reference point $x^A$ in the acceptance space was selected as the 75th percentile of each predictor (see second column in Table 3). Its PoD is $p(x^A) = 0.016$. We selected two points in the declined region: $x_1^D$ (third column) and $x_2^D$ (fifth column) with $p(x_1^D) = 0.294$ and $p(x_2^D) = 0.858$. Columns 4 and 6 in Table 3 provide the corresponding decompositions (or attributions) of the differences $[f(x_1^D) - f(x^A)]$ and $[f(x_2^D) - f(x^A)]$. They are positive for monotone increasing variables (note the value of x2 was flipped). They can be negative for non-monotone ones, and some of them are in Table 3. Values in parenthesis are percentages, component values divided by the total difference.

Consider Column 4 first. There is no difference in the values of x6, x7, and x8 (number of accounts and amount past due) between $x^A$ and $x_1^D$, so they do not contribute anything. The values of x3 and x5, which deal with utilization, are not very different, so their contributions are also small. The values of x1, average balance, are quite different for the declined and reference points, but we see from Figure 5 that this variable is not as important. So the contribution is relatively small (3.5%). On the other hand, the values of x2 in the first and third columns are quite different. Moreover, Figure 5 shows it is an important variable in the model. So the contribution is the largest, in absolute sense and percentage of about 60. The next biggest contribution is from x9, number of credit inquiries in the past 12 months, with a contribution of about 31%. Note that x9 has the second largest important score in the left panel of Figure 5.

| Predictors | $x^A$ | $x_1^D$ | M-NN Attributions for $x_1^D$ | $x_2^D$ | M-NN Attributions for $x_2^D$ |
|---|---|---|---|---|---|
| x1 = avg bal cards std | −0.006 | 0.674 | 0.112 (3.5%) | 0.519 | 0.028 (0.5%) |
| x2 = credit age std flip | −0.733 | 0.886 | **1.928 (59.5%)** | 0.431 | **1.565 (26.5%)** |
| x3 = pct over 50 uti | 0.518 | 0.531 | 0.001 (0.0%) | 0.522 | -0.001 (0.0%) |
| x4 = tot balance std | −0.001 | 0.562 | −0.008 (0.2%) | 1.968 | −0.201(−3.4%) |
| x5 = uti open card | 0.501 | 0.577 | 0.012 (0.4%) | 0.525 | −0.024(−0.4%) |
| x6 = num acc 30d past due 12 months | 0.000 | 0.000 | 0.0 (0.0%) | 4.000 | **1.850 (31.3%)** |
| x7 = num acc 60d past due 6 months | 0.000 | 0.000 | 0.0 (0.0%) | 2.000 | **0.984 (16.6%)** |
| x8 = tot amount currently past due std | 0.000 | 0.000 | 0.0 (0.0%) | 4.379 | **1.712 (28.9%)** |
| x9 = num credit inq 12 month | 0.000 | 3.000 | **1.010 (31.2%)** | 0.000 | 0.0 (0.0%) |
| x10 = num credit inq 24 month | 0.000 | 4.000 | 0.186 (5.7%) | 0.000 | 0.0 (0.0%) |
| $p(x)$ | 0.016 | 0.294 | | 0.858 | |
| $f(x) = logit(p(x))$ | −4.117 | −0.876 | | 1.797 | |

Table 4. AA explanations after collapsing predictors that are highly correlated

| **Groups of predictors** | M-NN Attributions for $x_1^D$ | M-NN Attributions for $x_2^D$ |
|---|---|---|
| balance | 0.126 (3.9%) | −0.328(−5.5%) |
| credit age std flip | **1.925 (59.4%)** | **1.785 (30.2%)** |
| utilization | 0.018 (0.5%) | −0.018(−0.3%) |
| num acc | 0.0 (0.0%) | **4.476 (75.7%)** |
| num inq | **1.173 (36.2%)** | 0.0 (0.0%) |

Surprisingly, the difference for x10 between the first and third columns is even larger, and yet the contribution is much smaller. This is due to the fact that x10 has the lowest score in the left panel of Figure 5. While this may be reasonable (credit inquiries in the past 12 months are more important than past 24 months), it may also be an artifact due to the high correlation between the two variables (see Figure 4). Therefore, we collapsed the highly correlated variables and considered the same subset of five variables on the right panel in Figure 5. We then computed the joint B-Shap explanation for these variables as described in Section 4.3. The second column in Table 4 provides these joint attributions. These five predictors are more interpretable in terms of measuring distinct underlying measures of creditworthiness, and it is easier to explain the results to applicants in terms of these. The final AA explanation for this case is simple: the two important predictors are credit age and number of inquiries, with the former much more important. Decision makers can now use the numerical results, as needed, to provide further information to the applicants.

The comparison between $x^A$ and $x_2^D$, the first and fourth columns, can be done in a similar manner. The difference in the defaults probabilities is now much larger. After decomposing the difference (column 5), we see that the biggest contributors are x2 (credit age) and x6, x7, and x8 (number of accounts and amount) past due. Variables x6-x8 are highly correlated (see Figure 4), so we again computed joint attributions for the variables after collapsing them into five. The third column in Table 4 shows the joint decomposition and attributions. Number of accounts/amount past due is by far the most important with credit age also as a significant contributor, and the explanation is again quite straightforward even though the underlying model is complex.

6. **Discussion: Shape Constraints with Credit-Decision Models**

In credit decision applications, the fitted model should satisfy certain desirable properties. We discuss two of them in this section,

### 6.1. Monotonicity

For the illustrative application in Section 5, subject-matter knowledge suggested that six of the 10 variables have a monotone relationship with PoD. Thus, we fitted a shape-constrained NN where the fitted model satisfied the monotone properties. We argue that this should be a general requirement for all models that deal with credit decisions, namely, the fitted model should satisfy shape constraints for relevant predictors. If we expect POD to be increasing or decreasing with an underlying measures of credit worthiness, then the fitted model should satisfy the property (see also Chen et al., 2022). Decisions based on models that violate such basic requirements will not provide reasonable AA explanations.

There are techniques and algorithms in the literature to fit monotone versions of ML algorithms. Liu et al. (1992) proposed an iterative NN algorithm that fits a penalized loss function (with a penalty for monotonicity), checks if the property is satisfied, increases the penalty if not, and iterates. This is the algorithm used in our example in Section 6. See also Cano et al. (2019) for other methods.

### 6.2. Continuity

Another desirable property is that the fitted POD should vary continuously as a function of continuous predictors. To illustrate this, consider the illustrative example in Figure 8 with two continuous predictors $x_1$ and $x_2$. Suppose we fit a discontinuous (piecewise constant) tree model which yields the fitted values and decision boundary in Figure 8. Here $C_j$'s are the constants corresponding the fitted PoDs, and the red line is the decision boundary separating the "accept" and "reject" regions. Take, for example, the red and green points $(x_1^D, x_2^D)$ and $(x_1^A, x_2^A)$, in Figure 8, corresponding to declined and accepted applications. Their PODs are $C_3$ and $C_5$, and this difference will remain the same regardless of how close the two points are to each other but remain in opposite sides of the decision region. There is no way one can rationalize such a decision rule and explain it to applicants.

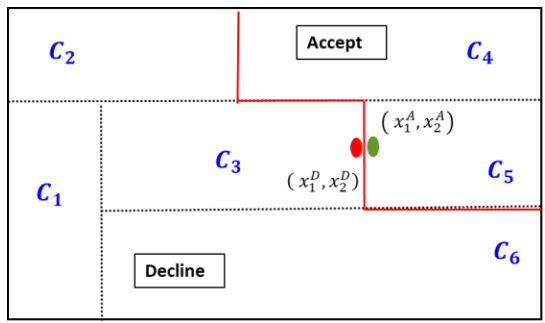

Figure 8: Graphical illustration for piecewise constant model

One of the commonly used approaches in credit-scoring models does in fact bin the continuous variables and fits piecewise constant models. This has been justified as a way to do feature engineering. As we argue above, such models cannot be justified if one has to provide meaning AA explanations. More broadly, tree-based ML algorithms such as Gradient Boosting and Random Forests all use piecewise constant fitting. Since they are ensemble models are based on many trees, the discontinuities may be small. However, they still pose conceptual challenges for AA explanation.

7. **Concluding Remarks**

This article proposes the use of B-Shap as a model-agnostic method for explaining adverse actions in credit decisions. It decomposes the difference of the PoD or its log-odds, evaluated at two points, into components corresponding to the different predictors. The method was developed from first principles. It is intuitive, so easy to understand and explain. The computations are easy for lower-order models. For arbitrary models, the computations are tractable even for dimensions as high as 20. When the number of predictors is larger, one would expect them to have significant correlations among subsets. In such case, as we have shown through our example, it is more meaningful to collapse them into fewer, distinct measures of credit-worthiness and do AA explanation in the reduced space.